\documentclass[10pt,twocolumn,letterpaper]{article}

\usepackage{cvpr}
\usepackage{times}
\usepackage{epsfig}
\usepackage{graphicx}
\usepackage{subfig}
\usepackage{tabularx}
\usepackage{algorithm}
\usepackage{algpseudocode}
\usepackage{amsmath}
\usepackage{amssymb}
\usepackage{gensymb}
\usepackage{multirow}
\usepackage{comment}
\usepackage{float}
\usepackage[normalem]{ulem}
\useunder{\uline}{\ul}{}

\usepackage[pagebackref=true,breaklinks=true,letterpaper=true,colorlinks,bookmarks=false]{hyperref}

\cvprfinalcopy 

\def\cvprPaperID{29} 

\ifcvprfinal\fi
\begin{document}

\title{A Novel Technique Combining Image Processing, Plant Development Properties, and the Hungarian Algorithm, to Improve Leaf Detection in Maize}

\author{
Nazifa Azam Khan$^1$, Oliver A.S. Lyon$^2$, Mark Eramian$^1$, Ian McQuillan$^1$\\
{\tt\small nazifa.khan@usask.ca}, {\tt\small oliver.lyon@queensu.ca}, {\tt\small mark.eramian@usask.ca}, {\tt\small mcquillan@cs.usask.ca}\\
$^1$Department of Computer Science, University of Saskatchewan, Saskatoon, SK, Canada\\
$^2$School of Computing, Queen's University, Kingston, ON, Canada}

\maketitle

\begin{abstract}
Manual determination of plant phenotypic properties such as plant architecture, growth, and health is very time consuming and sometimes destructive. Automatic image analysis has become a popular approach. This research aims to identify the position (and number) of leaves from a temporal sequence of high-quality indoor images consisting of multiple views, focussing in particular of images of maize. The procedure used a segmentation on the images, using the convex hull to pick the best view at each time step, followed by a  skeletonization of the corresponding image. To remove skeleton spurs, a discrete skeleton evolution pruning process was applied. Pre-existing statistics regarding maize development was incorporated to help differentiate between true leaves and false leaves. Furthermore, for each time step, leaves were matched to those of the previous and next three days using the graph-theoretic Hungarian algorithm. This matching algorithm can be used to both remove false positives, and also to predict true leaves, even if they were completely occluded from the image itself. The algorithm was evaluated using  an open dataset consisting of $13$ maize plants across 27 days from two different views. The total number of true leaves from the dataset was $1843$, and our proposed techniques detects a total of $1690$ leaves including $1674$ true leaves, and only $16$ false leaves, giving a recall of $90.8\%$, and a precision of $99.0\%$.
\end{abstract}

\section{Introduction}\label{intro}
Agriculture is the backbone of the world economy, and a significant number of countries' economies are highly dependent on it. Plant diseases, undesirable growth, nutritional deficiency, and disorder in plants not only affect the quality and quantity of agricultural profits, but also play a vital role in food crises. Thus, monitoring the condition of plants is a fundamental step in successful cultivation of crops and plant breeding. Indeed, plant breeding, with the assistance of high-throughput phenotyping, is helping to cultivate crops under extreme climate, and to create novel plant varieties \cite{article, high2019}. This can ultimately contribute towards a greater quantity and quality of food for feeding the ever-growing population. Until recently, the observation and analysis of plant growth, disease detection, and  phenotypic  properties, were done entirely manually by experts, in a time intensive, and largely intuitive fashion. Thus, the potential of using image processing in plant research to automate phenotypic inspection has long been recognised as an important step forward \cite{intro_1}. Now, the food industry ranks among the top industries using image processing  \cite{intro_2} to help evaluate food quality and consistency while eliminating the subjectivity of manual inspections \cite{intro_3}. 

Computer vision can be used to extract useful information from plant images \cite{plant}, and to identify phenotypic traits  throughout a plant's life \cite{crisp}. Various types of digital cameras are used to acquire  richer information about plants of interest \cite{translating, plant_model, plant_phenome}. Extracting meaningful phenotypes from plant image sequences is broadly classified into two categories: holistic and component-based \cite{automated}. Holistic plant phenotyping considers the whole plant as a single object and gives metrics that quantify the basic geometric properties of the plant (e.g. height, width, plant aspect ratio, etc). Component-based analysis tries to identify the specific distinguishing  components of a plant (leaves, stem, flower etc), their positions, and sizes \cite{unl}. 

\textbf{\textit{Problem overview:}} Our goal is to reconstruct and predict maize plant growth properties, topology, numbers and positions of leaves, and their emergence, from indoor time sequence plant images. Maize is a globally-grown annual cereal crop, and one  of the top three most important cereal crops in the world \cite{maize_2016, individual, agronomic}. Therefore, maize has a vital role to play in our agricultural economy, and automated prediction of maize plant growth, topology, components, disease, and architecture is important. 

Automatic determination of plant topology and architecture  is highly dependent on accurate plant skeletons.  Skeletons are a thin, sometimes one-pixel-wide, representation of any object that represents an object's topology; it is also often useful for feature extraction. After fifty years of research, there is still no perfect skeletonization algorithm for each individual area of application \cite{lam1992}.  Obtaining accurate plant skeletons from images is a difficult problem, as they are sensitive to small changes leading to extraneous branches, and  incorrectly joined segments (errors in topology) \cite{branches}. Extra branches, also called spurs, are especially common in plant skeletons and form due to noise in images \cite{cai}. Spurs are often incorrectly interpreted as leaves.

The complex geometry of plants, their thin structures, and missing information due to self-occlusion, make skeleton extraction and pruning extremely challenging tasks  \cite{ayan}.  Occlusion can occur frequently in 2D images, both partially, and totally. Partial occlusion occurs when a part of a component is occluded from an image, e.g. part of a leaf hiding its branching point. Total occlusion occurs when a leaf is totally obscured by other components of the plant. For total occlusion, there is no obvious way to tell from the image itself that a component is present.

\textbf{\textit{Contribution:}} This study proposes a novel technique  to improve detection of leaves and topology  in maize. The proposed method initially obtains the plant skeleton with image processing algorithms, and then it applies statistics regarding maize development available in literature to the skeleton to improve the predicted topology.   Lastly, the Hungarian algorithm is applied to match the leaves in each day's image with those in the previous and next day's images to match skeleton components between days. The Hungarian algorithm, also known as the Munkres algorithm, is an algorithm on weighted, undirected graphs that determines the one-to-one mapping between two given sets of vertices where the matched edges have the mathematically smallest combined weight \cite{hungarian}. Despite there being an exponential number of such mappings, the mathematically optimal solution can be found in polynomial time. This can be used to find the best matching between leaves in one image of a plant with those of the same plant on another day, ideally matching the same leaves together \cite{shortest}. This can both discard leaves detected from erroneous spurs, and also properly predict components even if they are completely occluded. In this way, the analyses are not completely dependent on the skeletonization techniques and pruning strategies. This contributes not only to leaf counting, but also to inference of plant topology.

While the analysis was carried out using images of maize, certain aspects of the analysis would be generalizable to time sequence images from other plant species. For example, the use of the Hungarian algorithm to match different components of the same plant between days is an interesting approach generally. Furthermore, the use of apriori knowledge regarding plant development in a given species can be used to override the classification of components identified by the computer vision algorithms. 

\section{Dataset}\label{data}
An open dataset was used from the University of Nebraska-Lincoln \cite{dataset}. This dataset, called UNL-CPPD-I,  has images of 13 different maize plants (with different genotypes). Plants were imaged once per day for 27 days using the visible light camera of the UNL Lemnatec Scanalyzer 3D high-throughput phenotyping facility \cite{unl}. Images were taken from two different orthogonal side views at 0 degrees and  90 degrees; denoted by  \textit{view-0 image},  and  \textit{view-90 image}, respectively. The 0-degree orientation is not always fixed across days, thus the best view for segmenting leaves differs from day to day even for the same plant. 

Maize has multiple stages of development; vegetative, transitional, reproductive, and seed \cite{bonnett}. All images in the dataset are only from the vegetative stage. During this stage, the tip of the main stem is short, leaves are arranged in an alternate phyllotaxy (each leaf develops on the opposite side of the previous leaf, forming a left-right alternating pattern), and leaves arise at a certain distance from the top of the  stem. A limited number of axillary buds can develop, but ears do not develop until further stages. Hence, at this stage, the topology is dominated by the alternating leaf pattern.

The dataset  also contains ground-truth annotated images with the visible leaves marked. Note that if a leaf is not visible in a given image, then it is not annotated in the ground-truth. This is immediately evident because the number of annotated leaves from the two views can differ substantially. While this is advantageous from the perspective of identifying leaves on an individual image, it does hinder the evaluation of leaf identification procedures that try to identify leaves even if they are occluded, which is our desired goal. 

The imaging started on October 10, 2015, 2 days after seed planting. 
The dataset contains 700 images. A detailed description about the imaging setup, dataset organization, and their genotypes is given in \cite{unl}.

\section{Methodology}\label{methods}
This section discusses the methods, and algorithm implementations.  Each phase is described in a subsection, and are image segmentation, view selection,  plant skeletonization, a threshold-based pruning method, spur removal based on statistics from literature on maize, and the use of the Hungarian matching algorithm to improve leaf counting.
Certain thresholds calculated within are appropriate for indoor time-sequence images of maize, and would likely need to be adjusted for other species and setups. However, the process used to derive the thresholds can be applied elsewhere, along with the aforementioned generalizable elements.

\subsection{Segmentation}\label{segmentation}
\begin{figure}[!htbp]
  \centering
  \subfloat[]{\includegraphics[width=0.24\linewidth, height=0.8in]{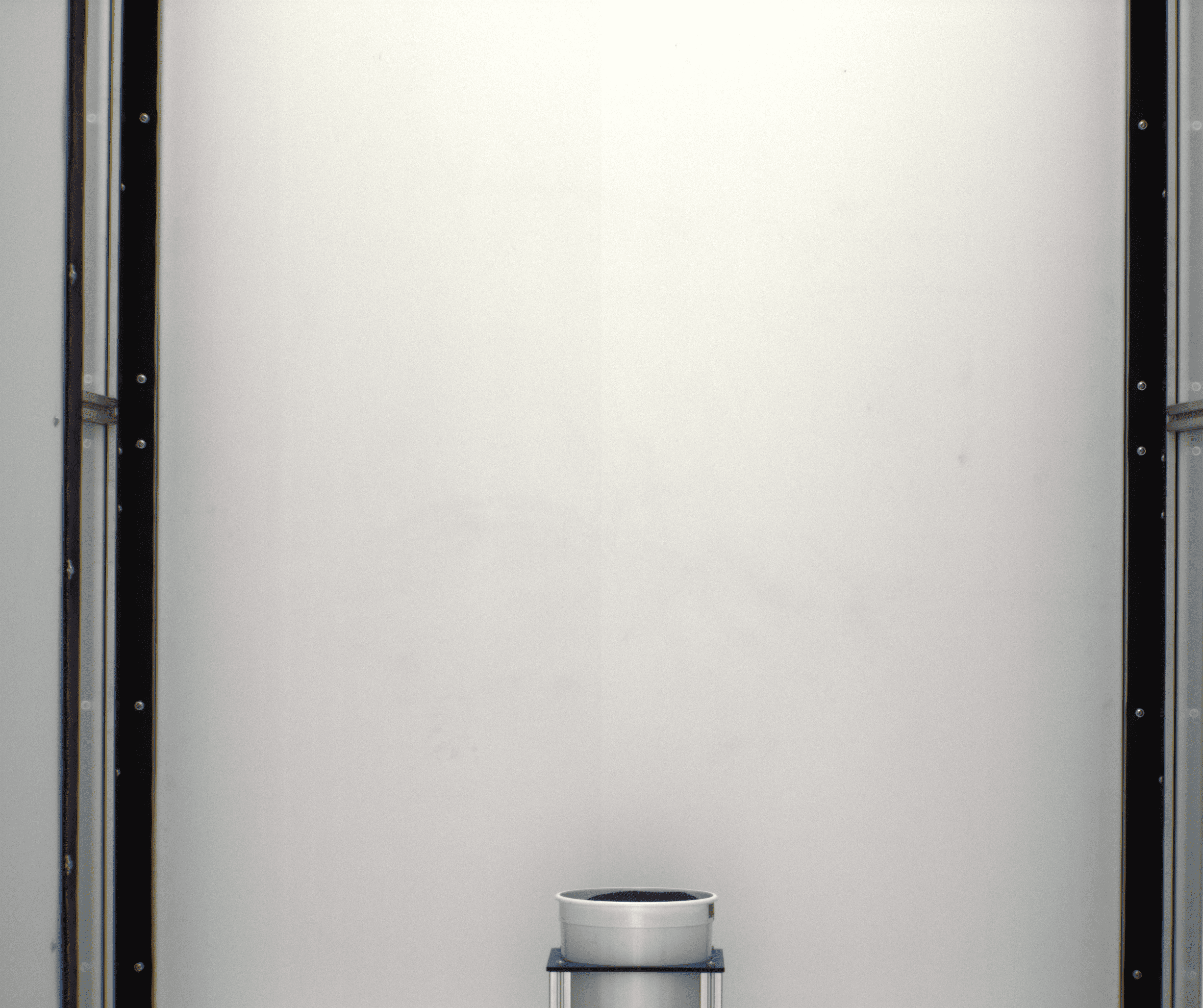}\label{fig_met_1}}
  \hfill
  \subfloat[]{\includegraphics[width=0.24\linewidth, height=0.8in]{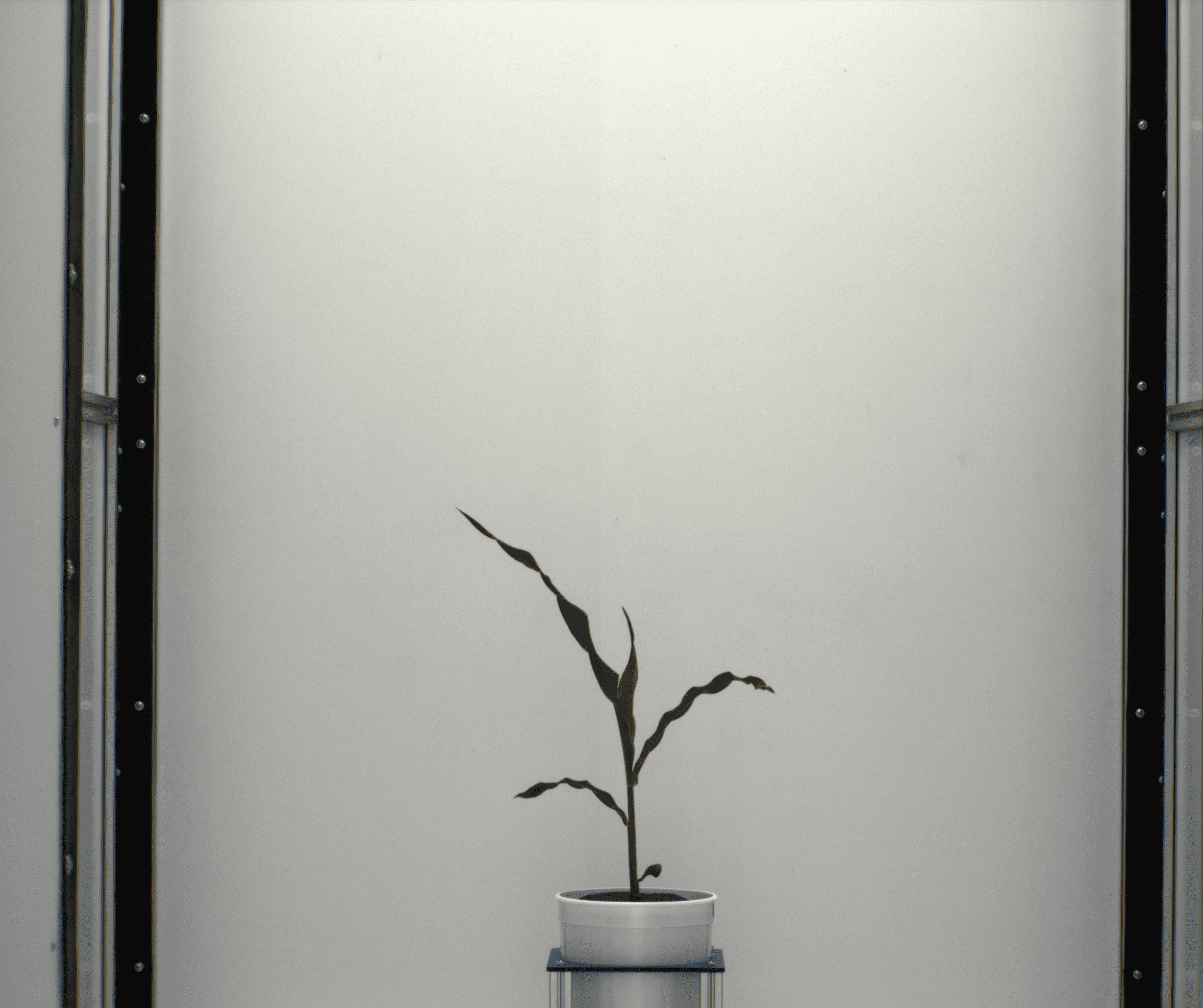}\label{fig_met_2}}
   \hfill
  \subfloat[]{\includegraphics[width=0.25\linewidth, height=0.8in]{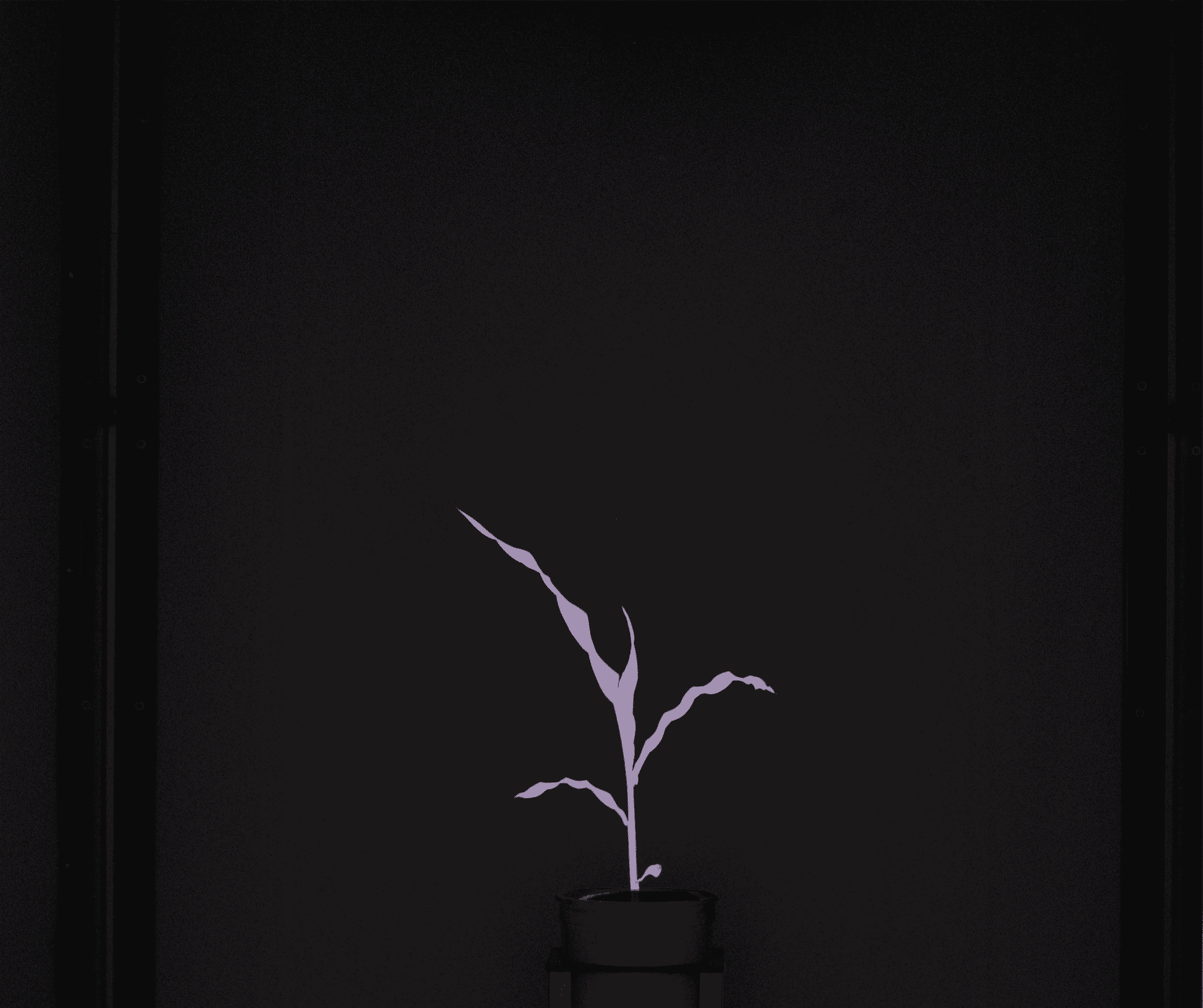}\label{fig_met_3}}
   \hfill
  \subfloat[]{\includegraphics[width=0.24\linewidth, height=0.8in]{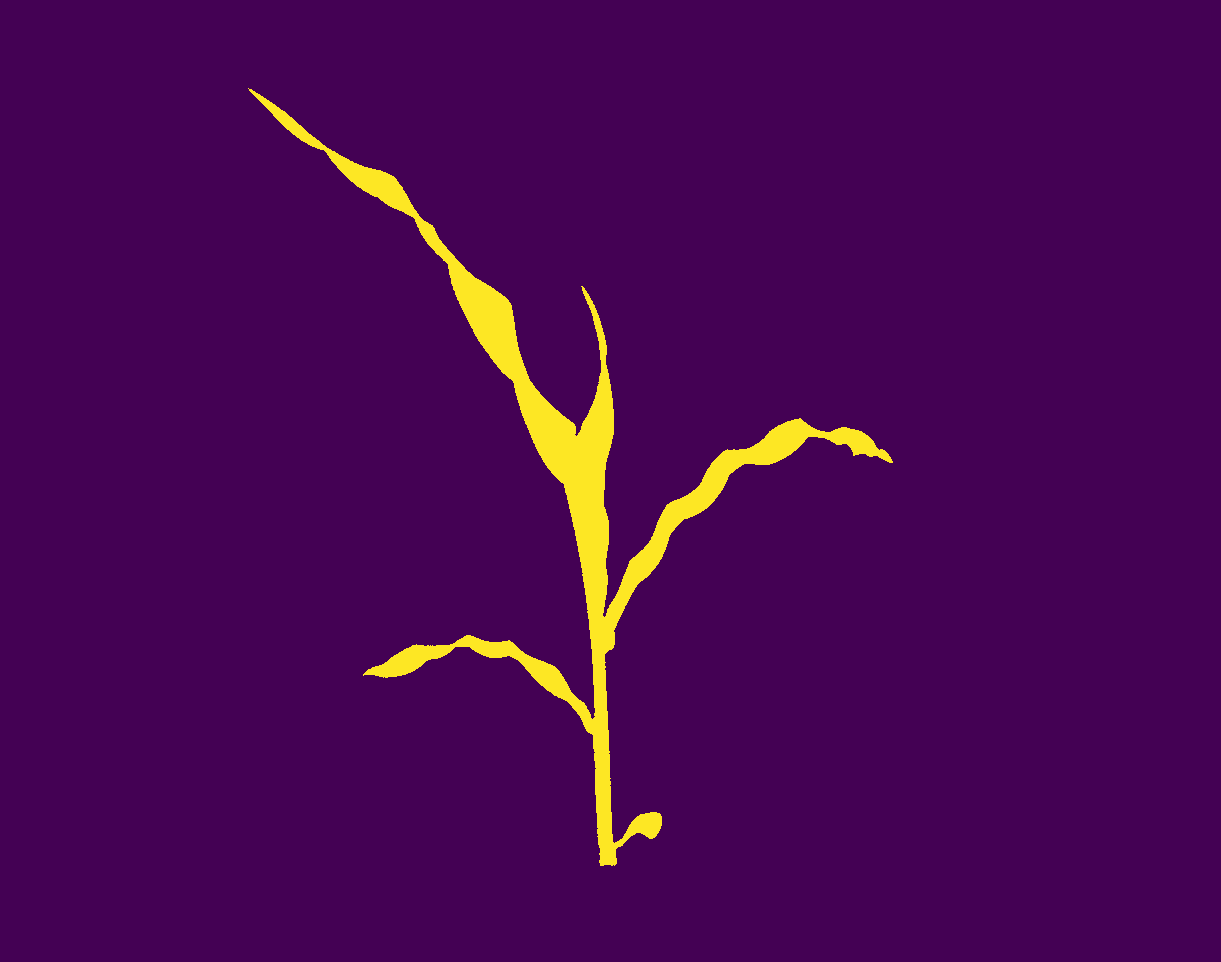}\label{fig_met_4}}
  \caption{(a) Background. (b) Plant\_001-9 at day 15 from view-90. (c) Plant foreground. (d) Segmented image.}
  \label{fig_met_5}
\end{figure}

The first step is obtaining the plant area from the available images with image segmentation techniques. Background subtraction was used to extract the foreground, which, in this case, is the plant itself. Background subtraction involves removing the background of the image,  which consists of the imaging chambers of the Lemnatec Scanalyzer 3D high-throughput plant phenotyping system. This has a fixed background that remains static over the period of interest for the image sequence \cite{unl} (Figure~\ref{fig_met_1}). Then, the Otsu thresholding algorithm  \cite{otsu} was used on  the grayscale image of the foreground image to obtain the segmented image. Figure~\ref{fig_met_5} shows an example of Plant\_001-9 at day 15 from view-90 (\ref{fig_met_2}), its foreground after background subtraction (\ref{fig_met_3}), and the resulting segmented plant image (\ref{fig_met_4}). 

\begin{figure}[!htbp]
  \centering
  \subfloat[]{\includegraphics[width=0.33\linewidth, height=0.9in]{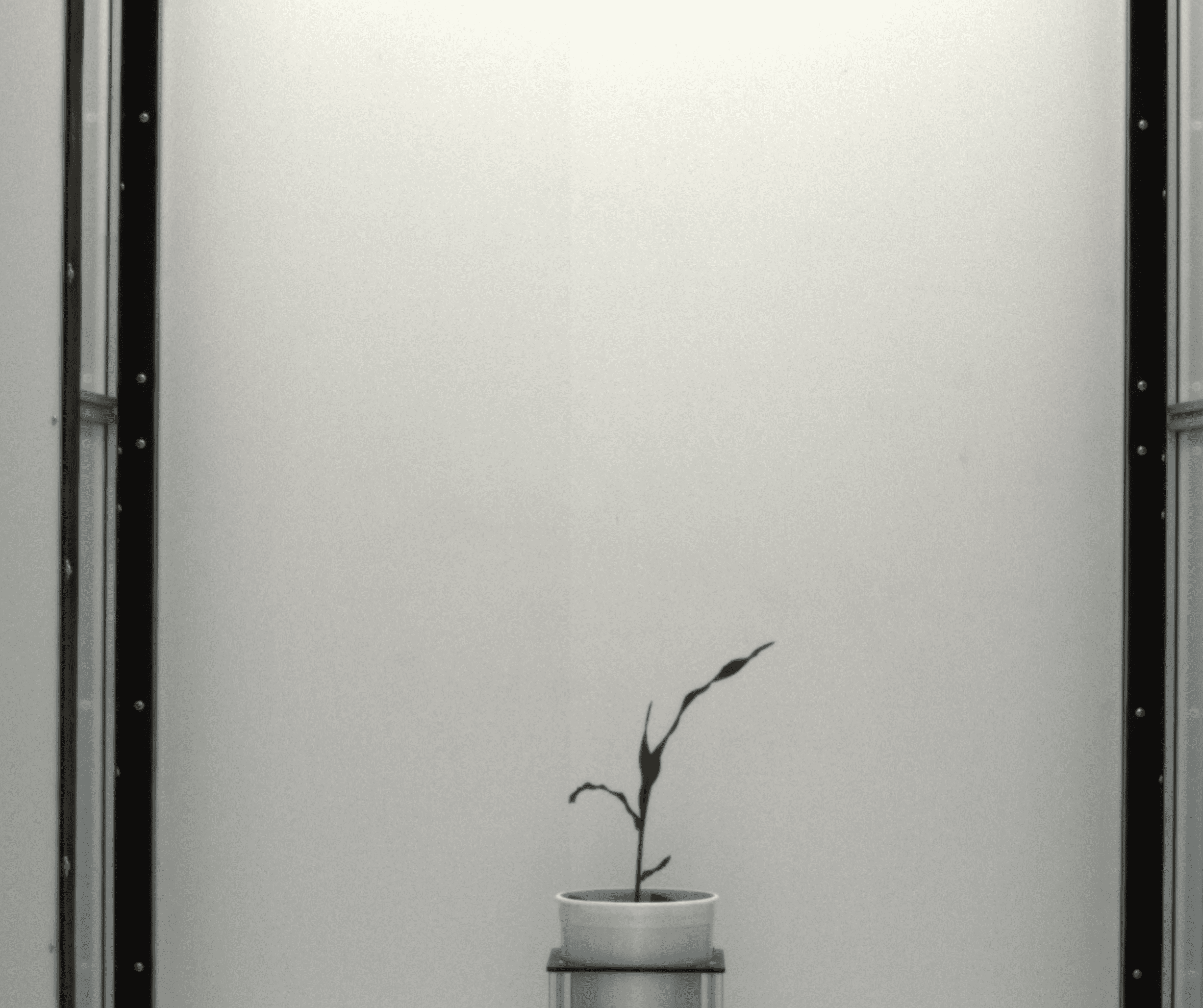}\label{fig_met_6}}
  \hfill
  \subfloat[]{\includegraphics[width=0.33\linewidth, height=0.9in]{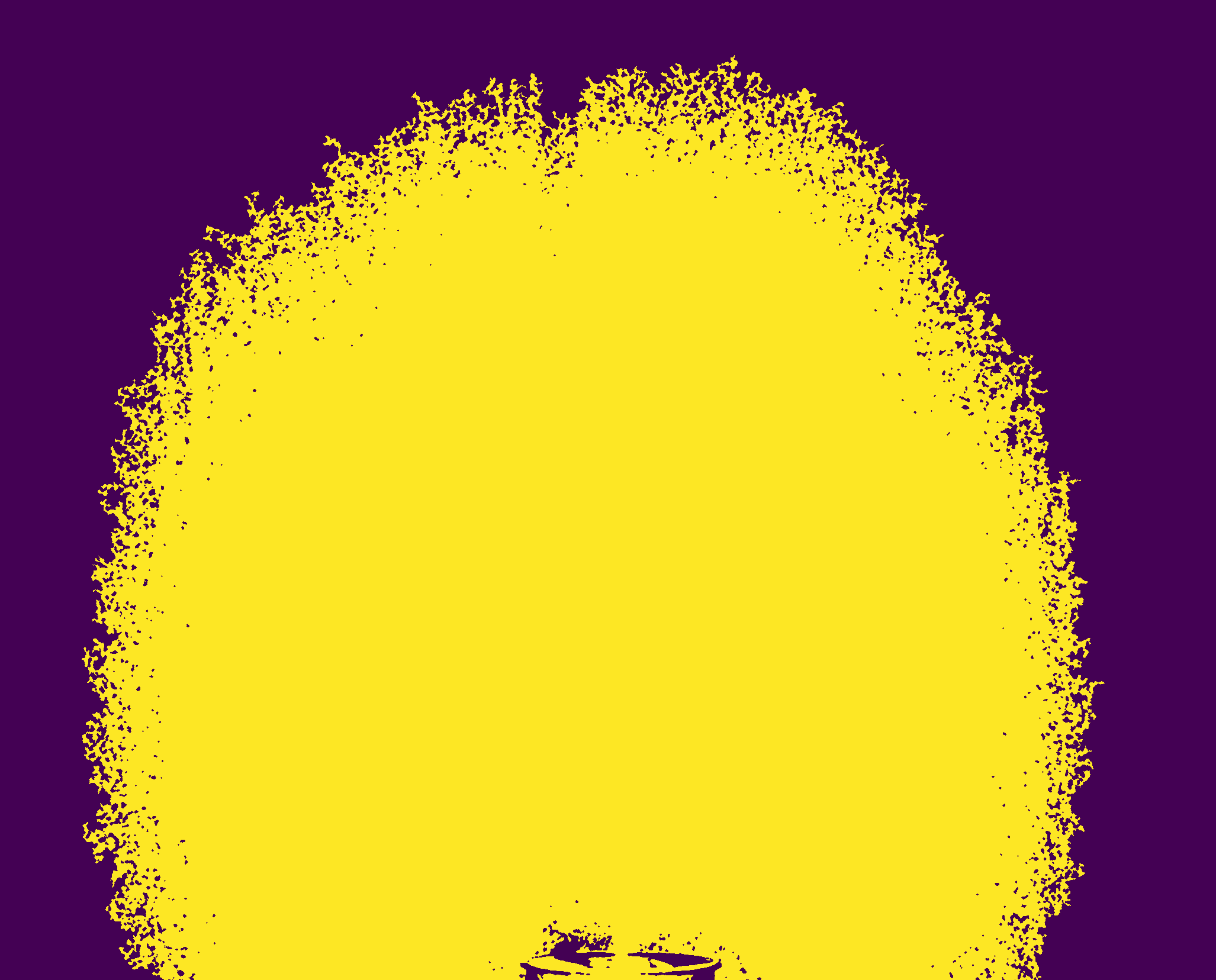}\label{fig_met_7}}
   \hfill
  \subfloat[]{\includegraphics[width=0.33\linewidth, height=0.9in]{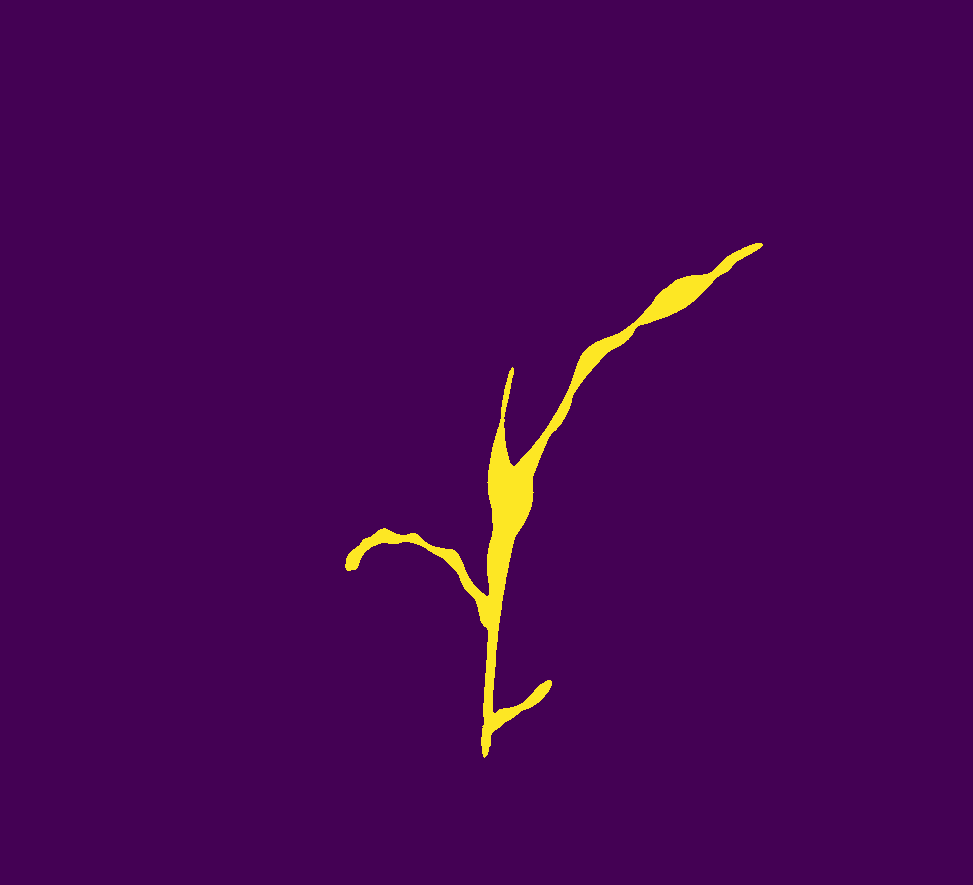}\label{fig_met_8}}
  \hfill
  \subfloat[]{\includegraphics[width=0.33\linewidth, height=0.9in]{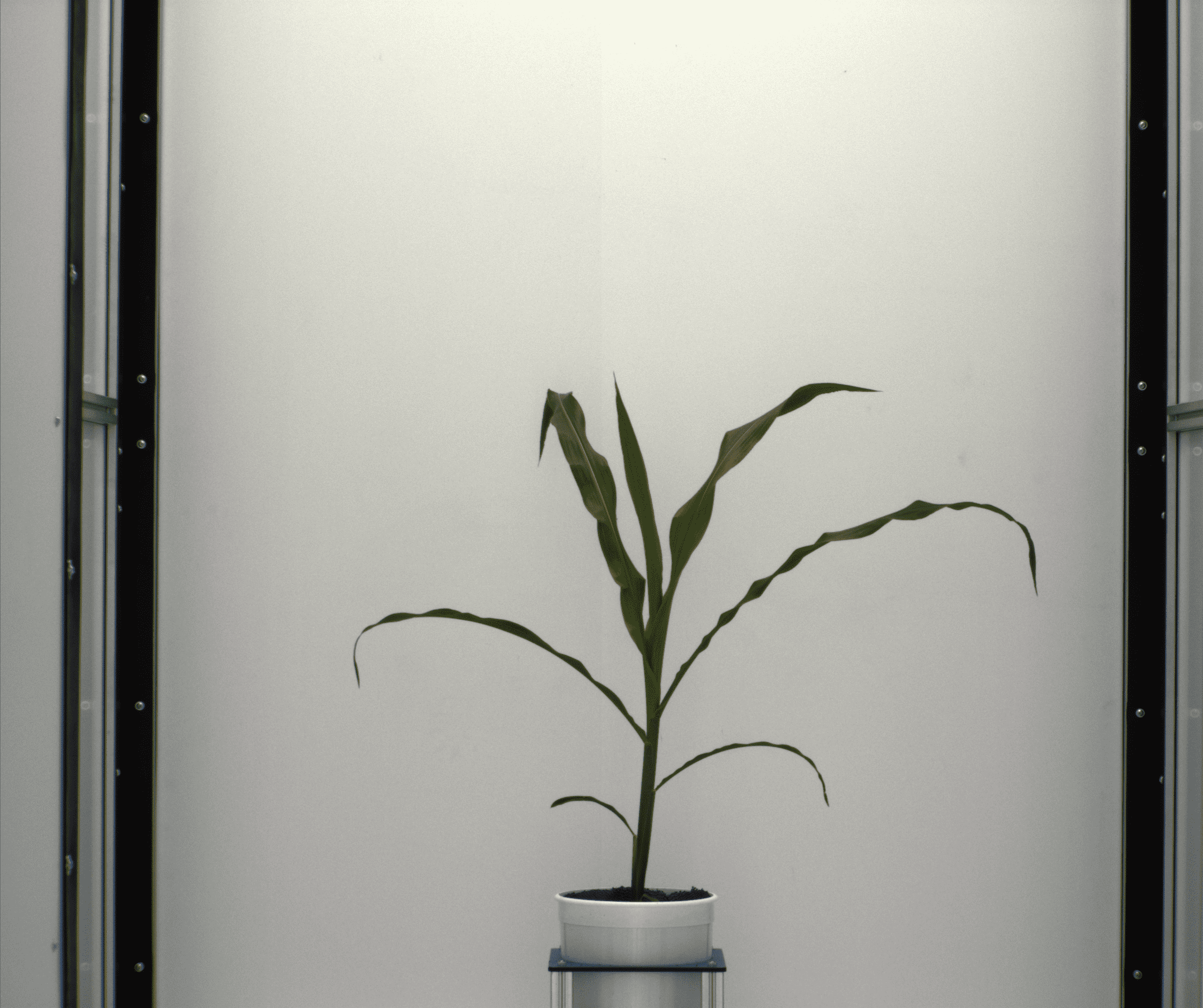}\label{fig_met_9}}
  \hfill
  \subfloat[]{\includegraphics[width=0.33\linewidth, height=0.9in]{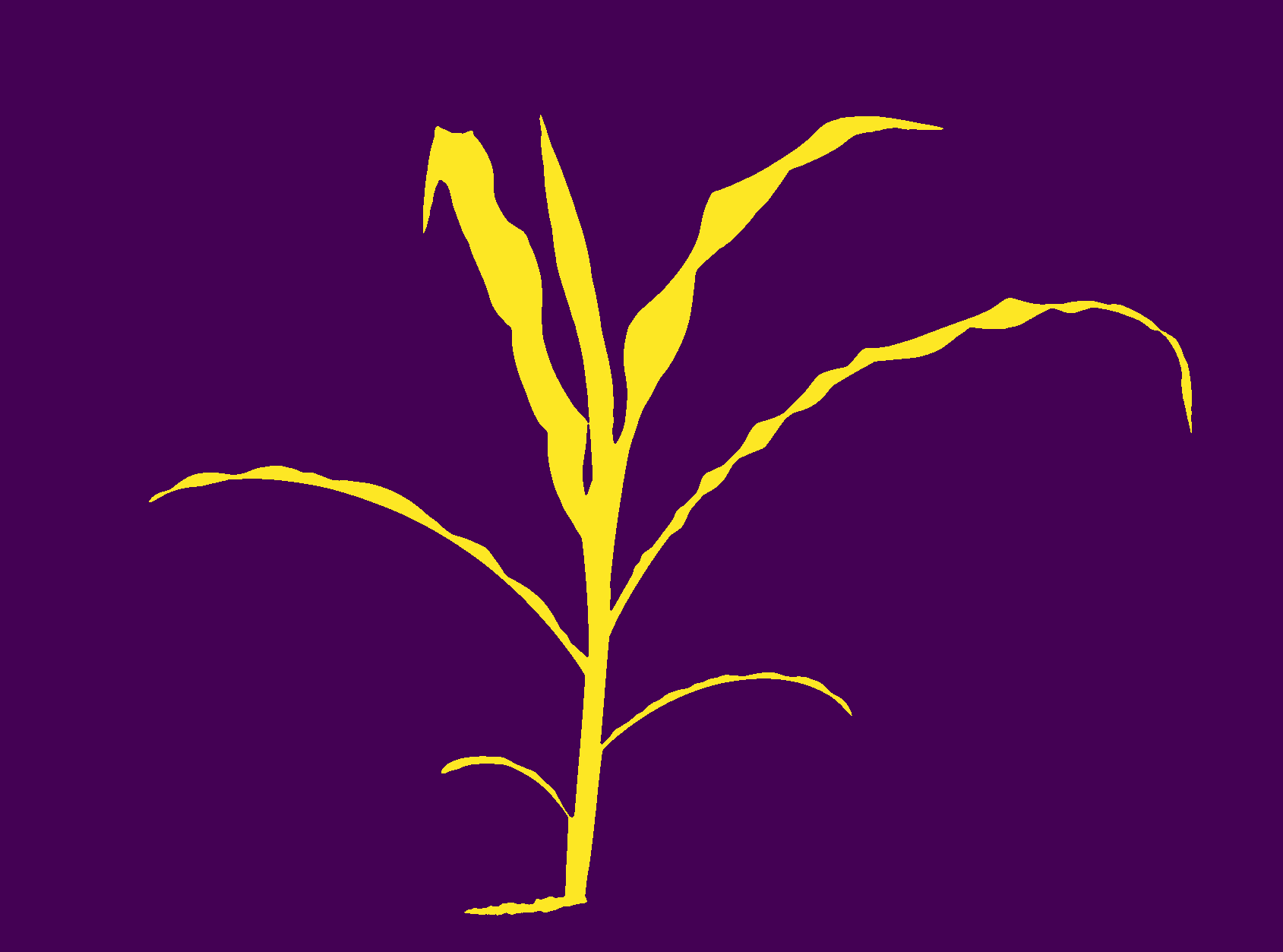}\label{fig_met_10}}
   \hfill
  \subfloat[]{\includegraphics[width=0.33\linewidth, height=0.9in]{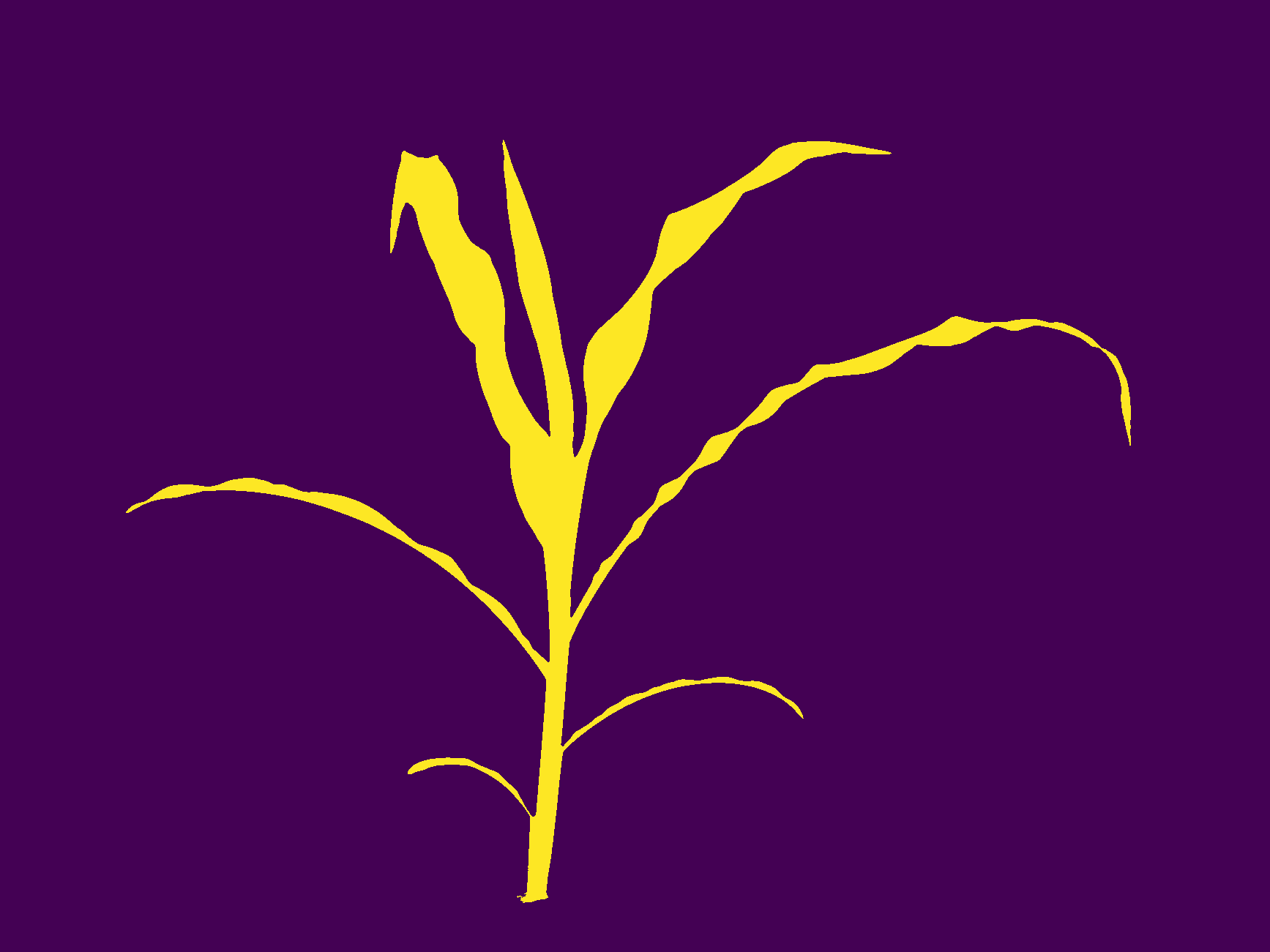}\label{fig_met_11}}
  \caption{(a) A plant image at day 11 \cite{unl}. (b) Segmentation with Otsu returned threshold smaller than $0.27$. (c) Segmentation with threshold value $0.27$. (d) A plant image at day 25 \cite{unl}. (e) The first thresholding with the tub pixels (the yellow line near the root is from the tub area). (f) After second thresholding, the tub pixels are removed.}
\end{figure}

Preliminary inspection of foreground image histograms showed that any threshold smaller than $0.27$ would label background pixels as foreground (Figure~\ref{fig_met_7}). However, for some images, the detected threshold was smaller than $0.27$ due to the light affecting the background. Therefore, the threshold used was the larger of $0.27$ and that detected by the Otsu algorithm. At this stage, there were some images where these thresholds were capturing some pixels from the plant tub (Figure~\ref{fig_met_10}). Hence, another level of thresholding was performed by calculating the excess green ($2G-R-B$) of the foreground image. The initially-thresholded pixels of the excess green image was thresholded again with a threshold value of the  maximum value between $0.1$, and the minimum value between Otsu returned threshold, and $0.5$; which is $t = \max(0.1, \min(t_o,0.5))$, where $t_o$ is the Otsu threshold, and $t$ is the final threshold value. Figure~\ref{fig_met_11} shows how the second level thresholding removed the tub pixels.

\subsection{View selection}\label{view}
\begin{figure}[!htbp]
  \centering
  \subfloat[]{\includegraphics[width=0.4\linewidth, height=1in]{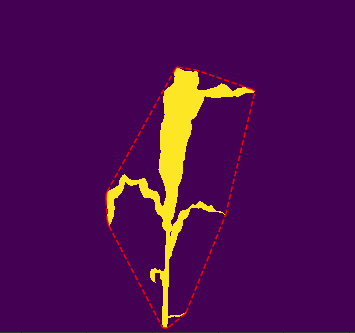}\label{fig_met_12}}
   \hfill
  \subfloat[]{\includegraphics[width=0.475\linewidth, height=1in]{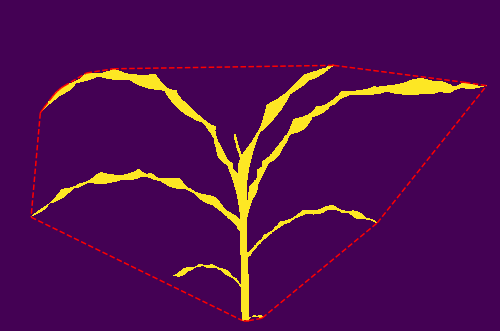}\label{fig_met_13}}
  \caption{(a) A segmented plant image at day 24 from view-0, and (b) view-90. The convex hull is outlined in red.}
\end{figure}

For each plant and day, a view selection process was applied to select the view (either 0 degrees or 90 degrees)  where the leaves, stem, and buds are most clearly visible. It is best to analyze the plant captured from the viewpoint at which as many leaves as possible are visible. Hence, we compute the area of the convex hulls of the binarized plant images of both views (a similar process was also used in \cite{unl}).  The view with the largest convex hull was selected. For example, Figures~\ref{fig_met_12} and \ref{fig_met_13} show the binary images of a maize plant on day 24, from both views. It is apparent that the area of the convex hull at side view 90 is higher.

\subsection{Skeletonization}\label{skel}
Skeletons are typically computed by either morphological thinning, computing the medial axis, geometric methods, or the fast marching distance transform. Morphological thinning takes a region, and gradually reduces the boundaries of that region until they are only separated by one pixel. The results of morphological thinning are similar to those of the medial axis transformation, which finds medial points by determining the set of points that are local maxima in terms of distance from the edge of the shape. Although these methods are straightforward, they require intensive heuristics to ensure connectivity of the skeleton in the case of complex dynamic structures such as plants \cite{unl}. 

After extensive preliminary testing, it was observed that different skeletonization algorithms work better in  specific ranges of days since emergence. This preliminary testing was measured based on the leaf count, spur count, and visually how accurately the skeleton branching points and tips are positioned. Branching points are the starting point of the leaf from the plant stem, and end-points are the leaf tips. 

Two different skeletonization methods were applied on different time intervals from emergence. The first skeletonization method is the fast parallel thinning algorithm  \cite{zhang}.  This approach works by making successive passes of the image and removing pixels on object borders; this continues until no more pixels can be removed. The image is correlated with a mask that assigns each pixel a number in the range $0 \dots 255$ corresponding to each possible pattern of its 8 neighbouring pixels. A lookup table is then used to assign each pixel a value of 0, 1, 2, or 3, which are selectively removed during the iterations \cite{zhang}.  This approach has the advantages of contour noise immunity and a good effect in thinning crossing lines \cite{chen}. Some of the earlier days' images have branches where lines representing leaves cross the stem. From a 1-pixel wide skeleton, a branching point was determined as a pixel with 3 or more neighbours, and a leaf end-point was the pixel with one neighbour. In earlier days' images, there are frequently overlapping lines in the skeleton, and the prediction needs to be able to properly classify the portions of the crossed lines \cite{saha}. In our testing, the fast parallel algorithm performs better at classifying these crossed lines. Hence, for skeletons from days 1 through 10 from emergence, this approach was applied.

However, this process causes numerous branching points in the skeleton near skeleton points that have more than three neighbours \cite{lam}, which occurs often in later days. Most of the images at later days have occlusions and curvatures in some leaves. Thus, an algorithm that is better in terms of noise sensitivity, and also preserves topologic and geometric connectivity would be better for these images.  Preliminary testing showed that the 3D medial surface/axis thinning algorithm performed well at resolving leaf occlusion and leaf curvature. Therefore, images from day 11 onwards were skeletonized by the 3D medial surface axis thinning algorithm \cite{lee}. This method uses an octree data structure to examine a $3\times3\times3$ neighbourhood of a pixel. The algorithm proceeds by iteratively sweeping, and removing pixels at each iteration until the image stops changing. Each iteration consists of two steps: first, a list of candidates for removal is assembled. Then pixels from this list are rechecked sequentially, to better preserve connectivity of the image \cite{lee}. The medial axis of an object is the set of all points that have more than one closest point on the object's boundary. It ultimately produces a 1-pixel wide skeleton preserving the  connectedness as the original object. 

\subsection{Skeleton pruning}\label{DSE}
The process of eliminating spurs to overcome skeleton instability is known as pruning \cite{pruning}. A fixed-threshold-based pruning method could be used on all maize images in an attempt to remove skeleton spurs \cite{cai}. However, this resulted in many false negatives. Therefore, a pruning method called discrete skeleton evolution  \cite{discrete} was applied on all of the plant skeletons. The fundamental theory of this process is to remove skeleton end-branches that have the smallest relevance for shape reconstruction. It calculates the relevance of branches as their contribution to shape reconstruction by calculating a weight for every edge between an end-point and a branching point iteratively, and any such edge having a weight less than a threshold is deleted. The weight is calculated with the following formula, $1 - (a_s - a_e)/a_s$; where $a_s$ is the current area of the skeleton, and $a_e$ is the area of the edge; the threshold of $0.005$ was used from \cite{discrete} (area is the number of pixels in that object). This is appropriate because a small weight $w_i$ indicates that the edge has a negligible influence on the skeleton reconstruction, and the skeleton can be reconstructed without this branch in nearly the same fashion as the reconstruction with it  \cite{discrete}.

\subsection{Eliminating skeleton spurs with heuristics and statistics regarding maize development}\label{pruning}
By attempting to detect and remove spurs using the thresholding technique from Section~\ref{DSE}, there is a risk that it will incorrectly identify some components as a spur and remove it. Therefore, the following are used to decide between a true leaf, and a spur.

\subsubsection{Removing one pixel long spurs} 
\begin{figure}[!ht]
  \centering
  \subfloat[]{\includegraphics[width=0.525\linewidth, height=1in]{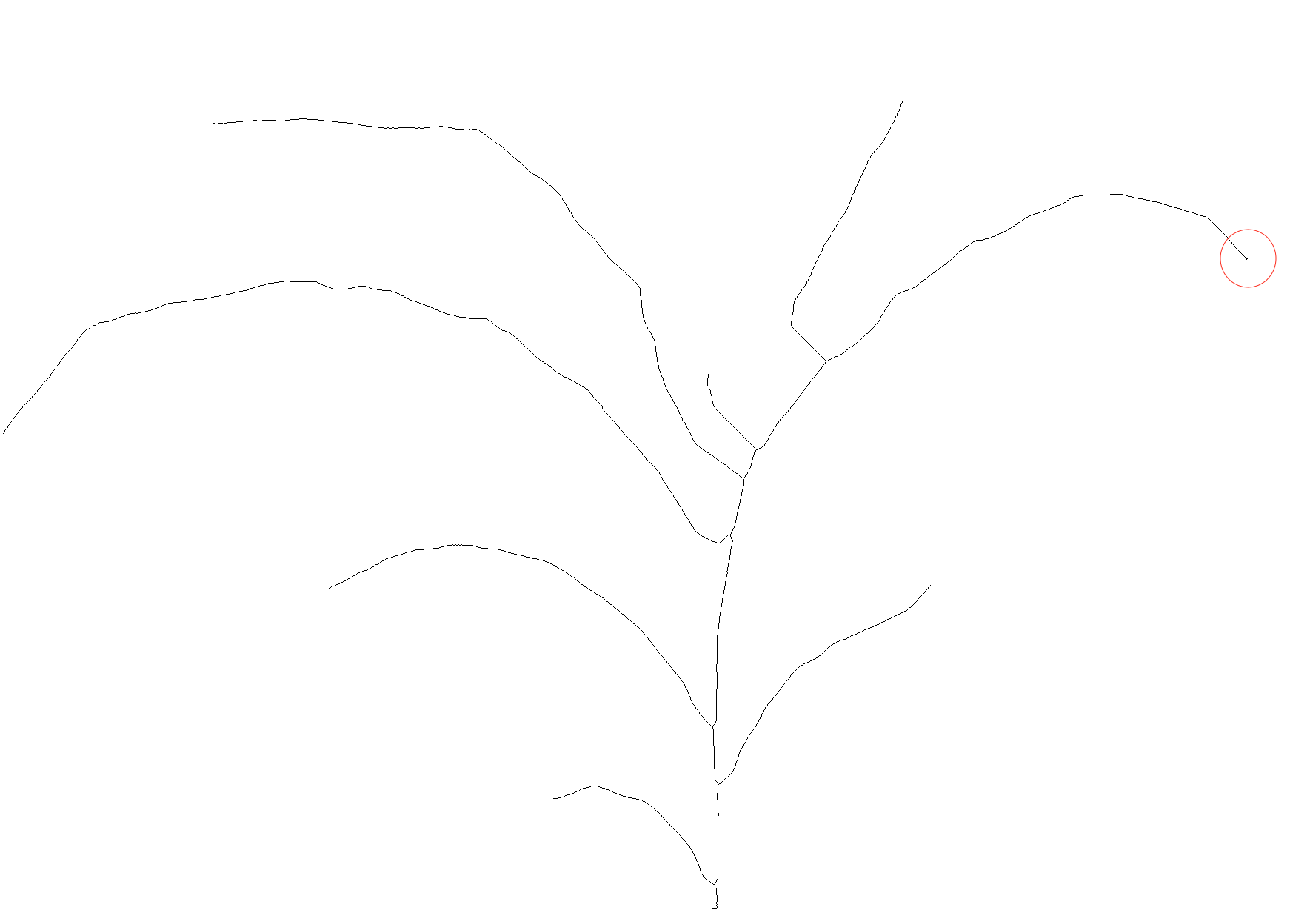}}
   \hfill
  \subfloat[]{\includegraphics[width=0.35\linewidth, height=0.75in]{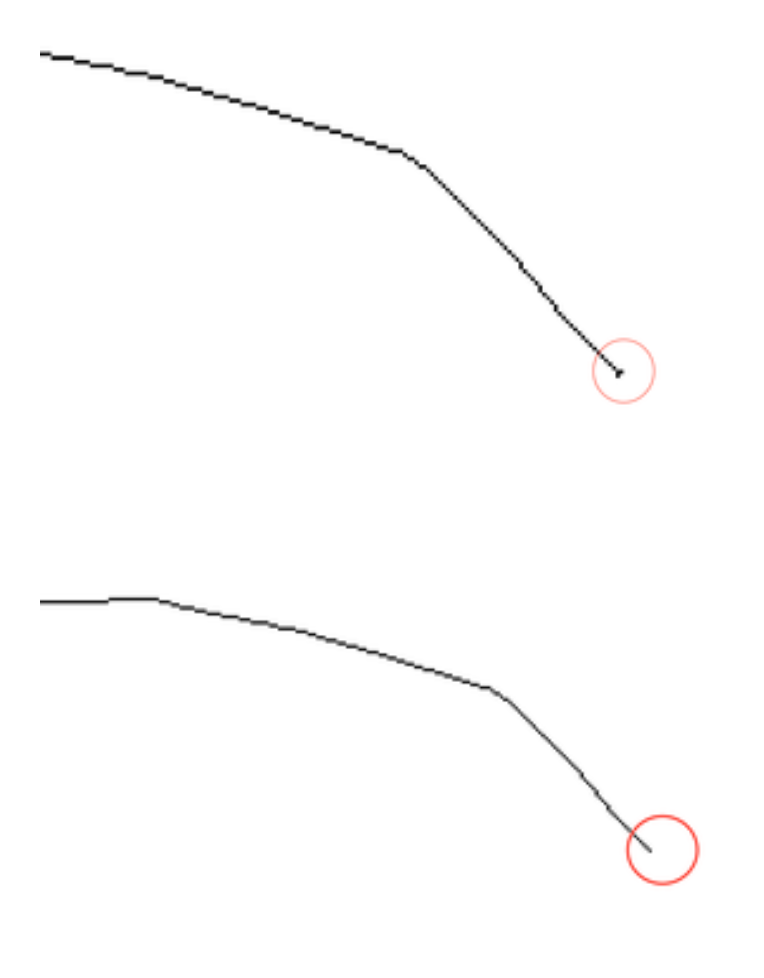}}
  \caption{(a) A plant skeleton image at day 26 having a 1 pixel long spur is circled in red, and (b) The spur is shown zoomed in circled in red (upper right), and after removing the spur (lower right).}
  \label{fig_met_14}
\end{figure}

A general threshold-based edge pruning was performed on the upper area. In the images, the root of the plants is always within $1700$ pixels from the bottom of the image. A $1$ pixel edge pruning was performed to remove any skeleton spurs that were $1$ pixel long above $1700$ pixels above the bottom. This was done because upper leaves would have emerged later on, and tended to be large (leaves produced early in development were more likely to be small). This threshold also helped remove spurs caused by leaf curvature. 
Figure~\ref{fig_met_14} shows such an example.

\subsubsection{Root area pruning with maize statistics} \label{step_2} 
\begin{figure}[!h]
  \centering
  \subfloat[]{\includegraphics[width=0.25\linewidth, height=0.6in]{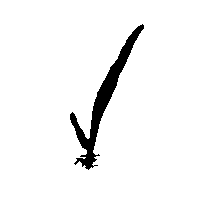}\label{fig_met_15}}
  \hfill
  \subfloat[]{\includegraphics[width=0.25\linewidth, height=0.6in]{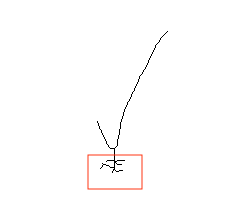}\label{fig_met_16}}
   \hfill
  \subfloat[]{\includegraphics[width=0.25\linewidth, height=0.6in]{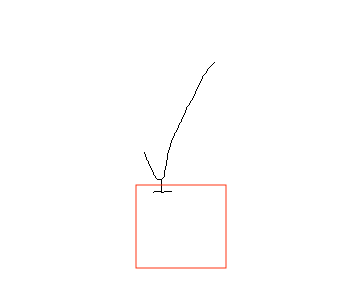}\label{fig_met_17}}
   \hfill
  \subfloat[]{\includegraphics[width=0.25\linewidth, height=0.6in]{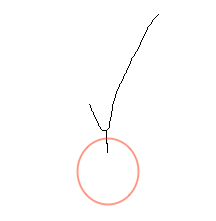}\label{fig_met_18}}
  \caption{(a) Segmentation of plant at day 5. (b) Skeleton of the same plant, and the spurs are shown within red rectangle (c) After root area cleaning with maize statistics (\ref{step_2}) (d) After root area cleaning by comparing last two consecutive branching point's position (\ref{step_3}).}
\end{figure}

There was a large number of spurs created adjacent to the tub edge and soil, resulting in uneven segmentation near the root of the plant. Thus, the next few pruning steps were performed to resolve this issue. This was based on existing information regarding the collar (a spot on the stem from where leaves emerge); the third leaf collar of maize plants usually becomes visible  approximately between 10 to 14 days after emergence \cite{corn}. Hence, we calculated the number of branching points, starting from the topmost (maize has an apical structure \cite{irish}, which means the new leaves should only emerge towards the top) for each plant image up to day 10, and everything below the fourth branching point was removed. However, even though the rule suggests there would be at most three branching points, four was chosen to be safe. Figure~\ref{fig_met_16} shows how this pruning step removed a number of spurs around the root of Figure~\ref{fig_met_17}. 

\subsubsection{Root area pruning by comparing last two consecutive branching point's position} \label{step_3} 
However, there were still some spurs near the root (Figure~\ref{fig_met_17}). After the above pruning steps, up to four branching points were still possible up until day 10. To remove putative spurs that remain, if their distance between the lowest two branching points is small (smaller than 10 pixels), then it is unlikely for them to be real leaves following alternate phyllotaxy. Hence, the lowest branching point was removed (Figure~\ref{fig_met_18}). This was also applied on plants up to day 10.

\subsubsection{Removing tub edge} 
\begin{figure}[!h]
  \centering
  \subfloat[]{\includegraphics[width=0.225\linewidth, height=0.7in]{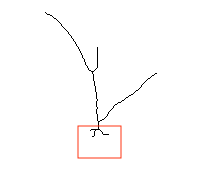}\label{fig_met_19}}
  \hfill
  \subfloat[]{\includegraphics[width=0.225\linewidth, height=0.7in]{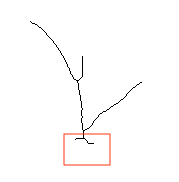}\label{fig_met_20}}
   \hfill
  \subfloat[]{\includegraphics[width=0.225\linewidth, height=0.7in]{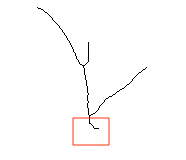}\label{fig_met_21}}
   \hfill
  \subfloat[]{\includegraphics[width=0.225\linewidth, height=0.7in]{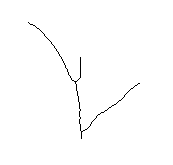}\label{fig_met_22}}
  \caption{(a) Skeleton of a plant at day 5. (b) Skeleton of the same plant, with spurs shown within the red rectangle (c) After root area cleaning with maize statistics and comparing last two consecutive branching point's position (\ref{step_3}). (d) After applying conditions to remove the tub edge. }
\end{figure}

Some images had spurs near the root even after applying the above pruning steps, for example, Figure~\ref{fig_met_21} (and on images after day 10). Hence, for all of the images, the $x$-distance, and $y$-distance\footnote{In images, the upper leftmost pixel is $(0,0)$, the $x$-axis is the height, and $y$-axis is the length.} between the lowest branching point, and the lowest end-point of the skeleton were calculated, which could either be an end-point of a leaf, or an end-point of an unwanted edge or spur, or possibly the root. If the $x$-distance is less than or equal to the $y$-distance, then possibly it is the root, or a leaf end-point; otherwise it might be a spur, and was deleted. This is because the lowest end-point should be on the stem, and a larger $y$-distance would be representative of moving horizontally from the stem without branching, a likely sign that it was created by the soil, or the tub edge. Figure~\ref{fig_met_22} shows how this pruning strategy removed the spur nearest the root.

\subsubsection{Remove root branches created due to non-smooth segmentation boundary} 
\begin{figure}[!h]
  \centering
  \subfloat[]{\includegraphics[width=0.25\linewidth, height=0.6in]{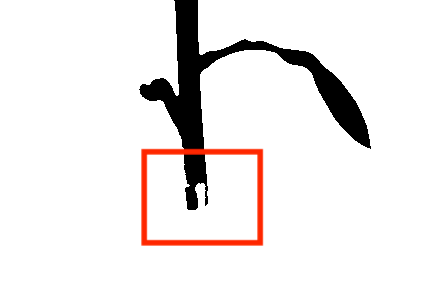}\label{fig_met_23}}
  \hfill
  \subfloat[]{\includegraphics[width=0.25\linewidth, height=0.6in]{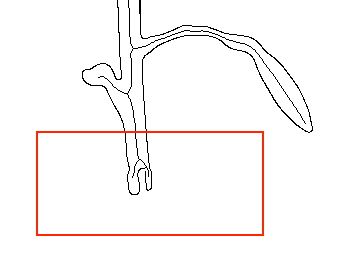}\label{fig_met_25}}
   \hfill
  \subfloat[]{\includegraphics[width=0.25\linewidth, height=0.6in]{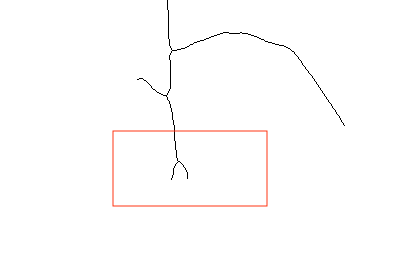}\label{fig_met_26}}
     \hfill
  \subfloat[]{\includegraphics[width=0.25\linewidth, height=0.6in]{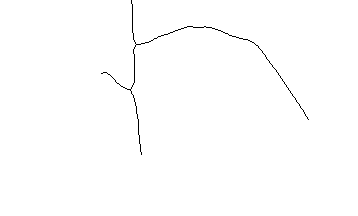}\label{fig_met_27}}
  \caption{(a) Segmentation of a plant at day 25. (b) Skeleton of the same plant overlaid inside plant boundary, and the spurs near  the root area are shown within red rectangle. (c) Plant skeleton without the plant boundary, and the spurs are within the red rectangle. (d) After removing root branches.}
\end{figure}

Lastly, while analyzing false leaves, it was noticed that uneven boundaries near the root of the segmented plant image causes false positives. Figure~\ref{fig_met_23} shows an example of such a scenario. The main challenge here was to decide between a bent dying leaf, and a spur. To identify and remove spurs such as this, the boundary of the segmented plant is compared to the lowest end-point of the first detected leaf  to assess whether it is in the middle of the boundary or not (Figure~\ref{fig_met_25}). However, there might be cases where the lowest edge is a bent leaf. To ensure that any true leaf is not deleted, the angles between the potential leaf and stem was calculated, and an angle threshold was used to make the decision. Figure~\ref{fig_met_26}, and \ref{fig_met_27} show how the angle between the potential leaf and stem play a role in deciding between a leaf and a spur. This pruning step was applied on plants at day 15 and later, where this scenario was more prominent.

\subsubsection{Removal of root spurs by comparing the lowest branching point with the lowest skeleton point} 
\begin{figure}[!h]
  \centering
  \subfloat[]{\includegraphics[width=0.225\linewidth, height=.7in]{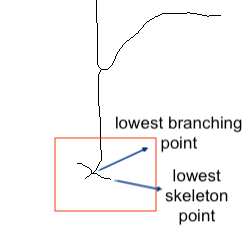}\label{fig_met_28}}
     \hfill
  \subfloat[]{\includegraphics[width=0.225\linewidth, height=.7in]{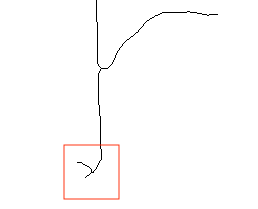}\label{fig_met_29}}
     \hfill
  \subfloat[]{\includegraphics[width=0.225\linewidth, height=.7in]{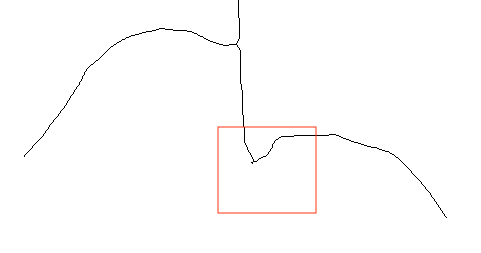}\label{fig_met_30}}
     \hfill
  \subfloat[]{\includegraphics[width=0.225\linewidth, height=.7in]{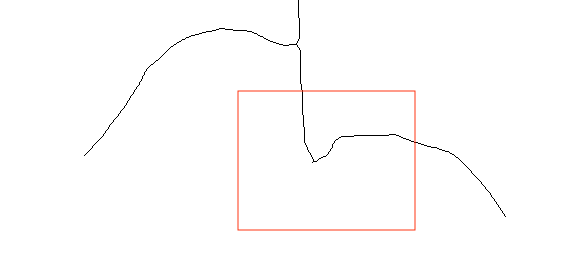}\label{fig_met_31}}
  \caption{(a) A plant skeleton at day 17, and the spurs are shown within red rectangle (the lowest branching point and the lowest pixel point of the skeleton are shown with arrow). (b) After pruning (c) A plant skeleton at day 26, and the spurs are shown within red rectangle. d) After pruning.}
\end{figure}

There were some skeletons that had a true leaf and a spur connected with the same branching point; this was common at the lowest branching point. In such a case, there were three branching points (Figure~\ref{fig_met_28}). We classify one as a continuation of the stem, one a leaf, and one a spur. 
A heuristic was used that looked at differences in both $x$ and $y$ coordinates of each end-point with the branching point. The stem has the smallest difference in $y$-coordinate. Between the two remaining segments, a decision is made based on the length of the segment with small and low segments being preferred as the spur. In Figure~\ref{fig_met_28}, after stem identification, the spur is associated with the lower of the two segments. In Figure~\ref{fig_met_30}, the lowest end-point is that of the leaf, but the spur is chosen to be a short segment.

\subsection{Growth properties, and leaf matching with the Hungarian matching algorithm}\label{matching}
Despite the fact that the pruning steps of Section~\ref{pruning} are largely helpful, there are some real leaves that are being improperly disregarded by them.  Also, there were still some false positive skeleton spurs present in some plant skeleton images. This section describes techniques to remove some more skeleton spurs, and also detect some true leaves that were not detectable with the image processing techniques.

It is helpful to understand some statistics regarding maize development. It has been found that until the tenth-leaf stage (meaning ten leaves with a collar are visible) the rate of leaf development is approximately 2 to 3 days per additional leaf \cite{corn_stat}. Thus, for each plant, we compared the number of detected leaves on each image between days.  It would be better to not compare the number of leaves, but to match leaves between days. But, the view selected is not the same for a plant across all days, hence matching in these cases is not straightforward. Hence, as a first pass, the number of  leaves between days was compared, and then if the numbers differed as described below, and the views were the same, then a matching algorithm was used.

\begin{figure}[!htbp]
  \centering
  \subfloat[]{\includegraphics[width=0.245\linewidth, height=0.95in]{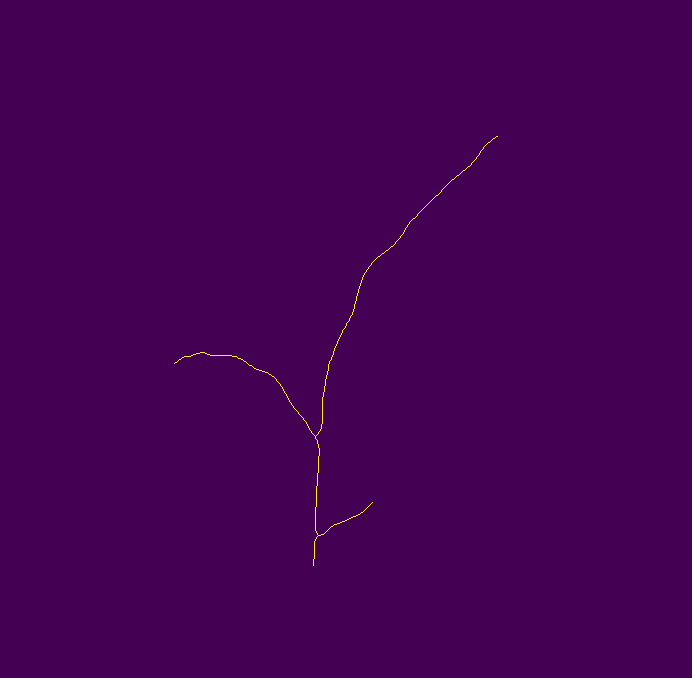}\label{fig_met_34}}
  \hfill
  \subfloat[]{\includegraphics[width=0.245\linewidth, height=0.95in]{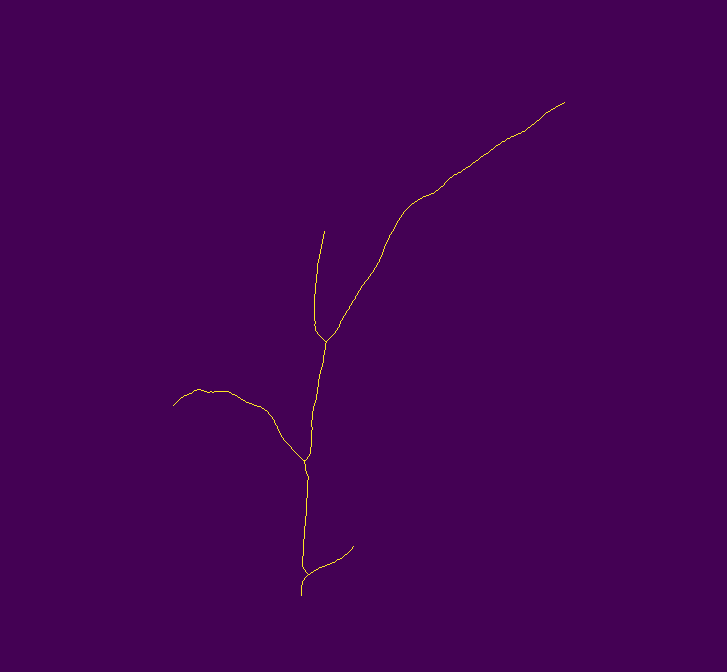}\label{fig_met_35}}
   \hfill
  \subfloat[]{\includegraphics[width=0.245\linewidth, height=0.95in]{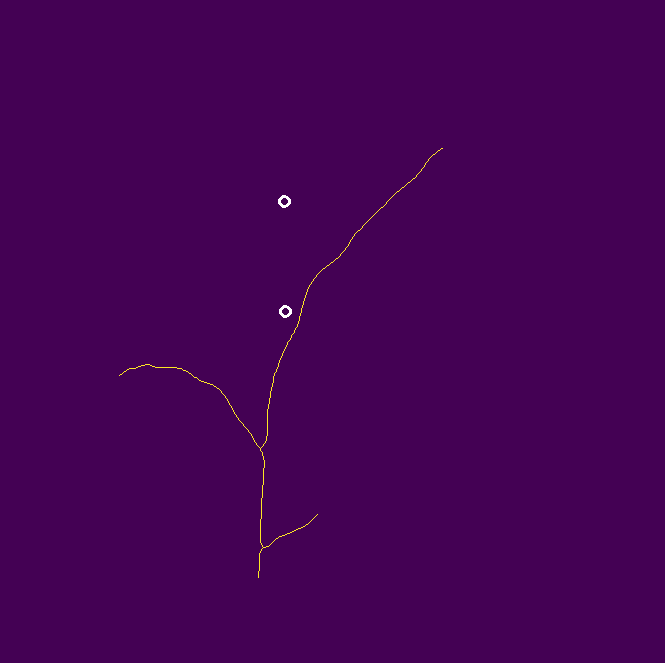}\label{fig_met_36}}
    \hfill
   \subfloat[]{\includegraphics[width=0.245\linewidth, height=0.95in]{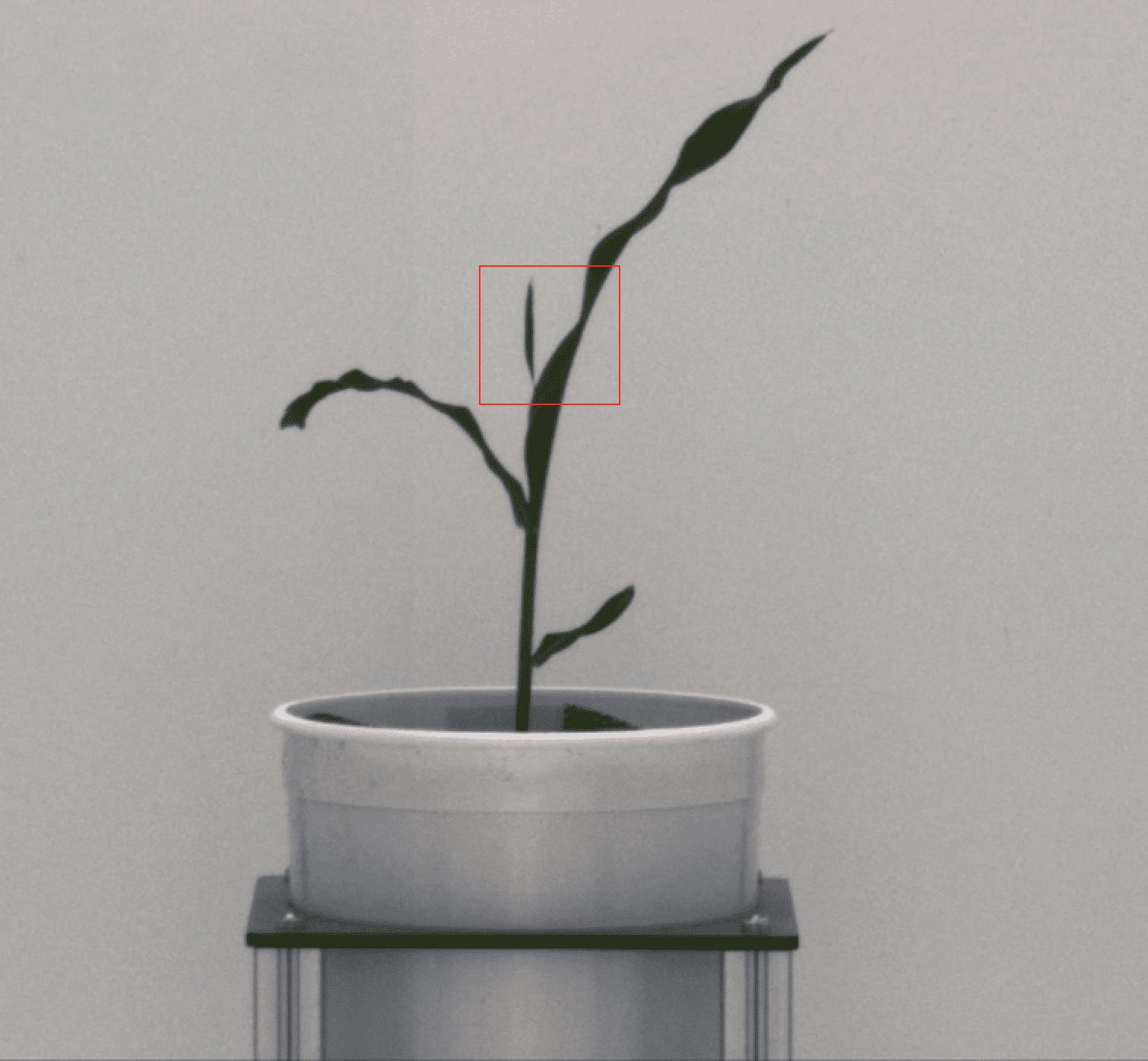}\label{fig_met_39}}
  \caption{(a) A plant skeleton at day 10 missing a leaf. (b) Skeleton of the same plant at day 11 having the leaf that is missed in day 10 skeleton. (c) The Hungarian matching algorithm successfully detects the missing leaf. (d) The plant image at day 10, and the leaf that was missing shown in red.}
  
\end{figure}

\begin{figure}[!htbp]
  \centering
  \subfloat[]{\includegraphics[width=0.245\linewidth, height=1in]{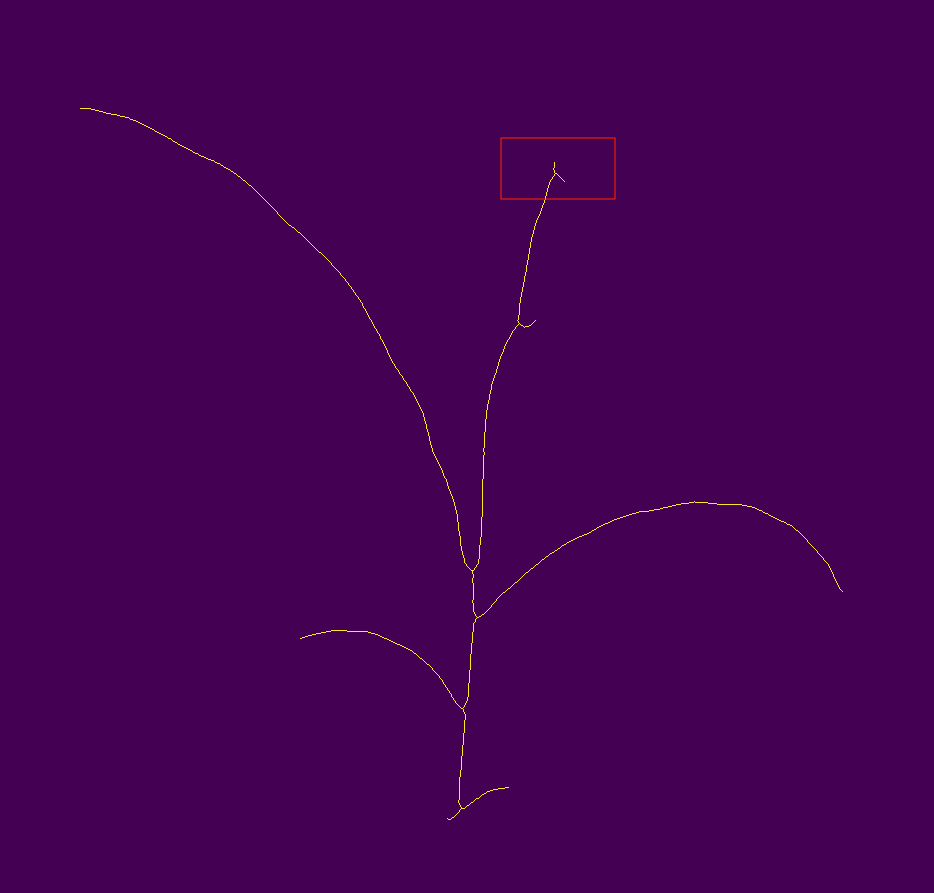}}
  \hfill
  \subfloat[]{\includegraphics[width=0.245\linewidth, height=1in]{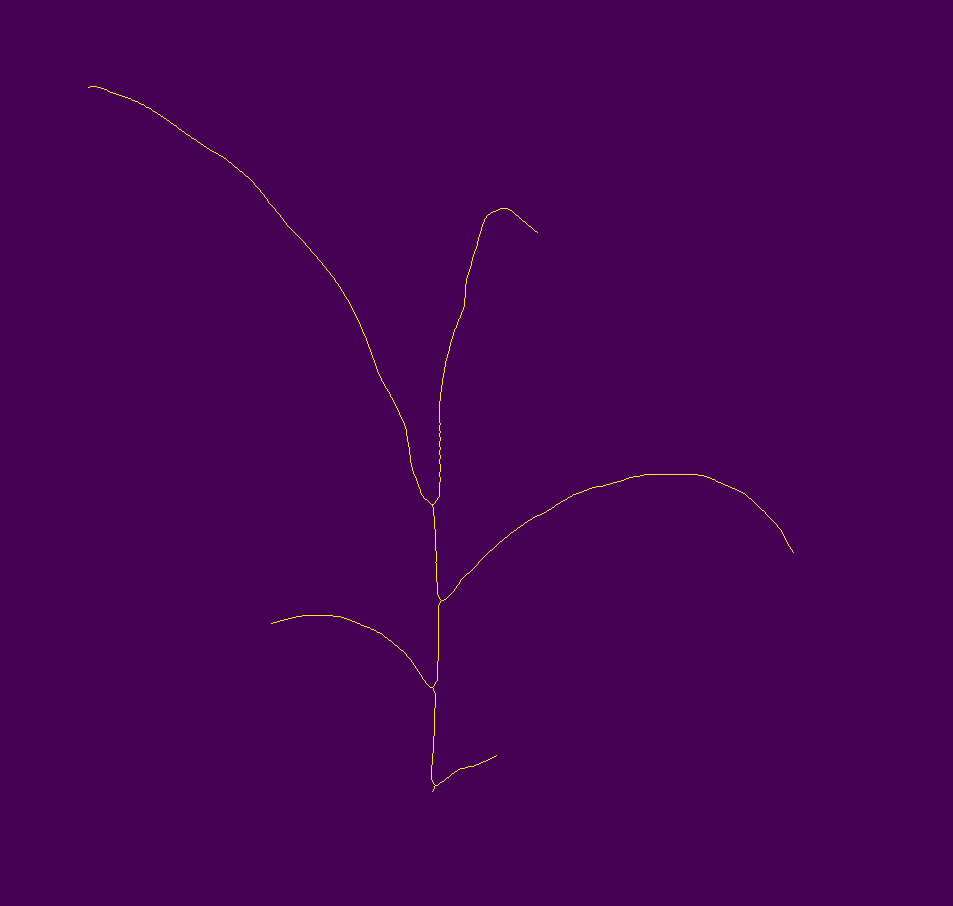}}
   \hfill
  \subfloat[]{\includegraphics[width=0.245\linewidth, height=1in]{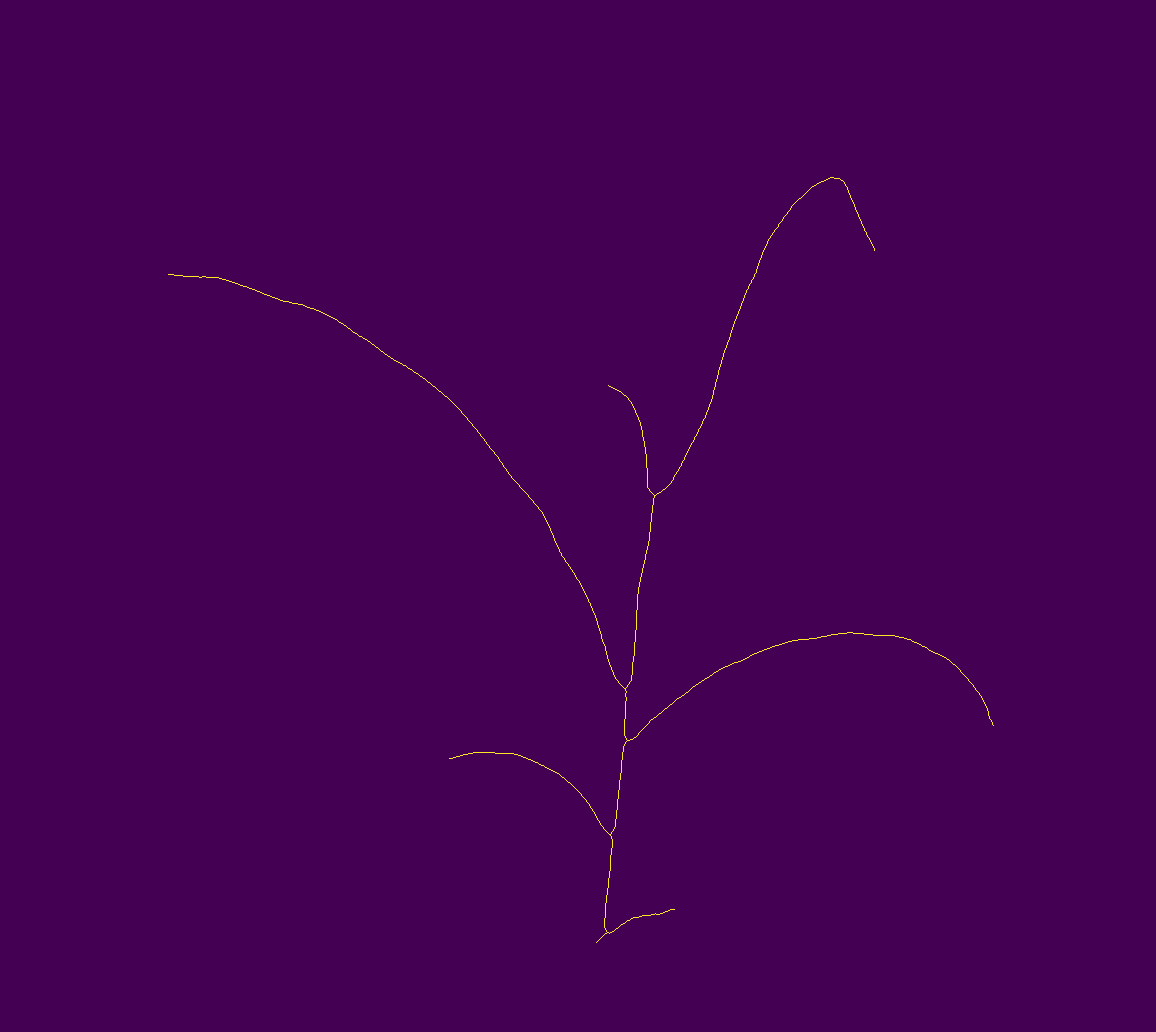}}
   \hfill
  \subfloat[]{\includegraphics[width=0.245\linewidth, height=1in]{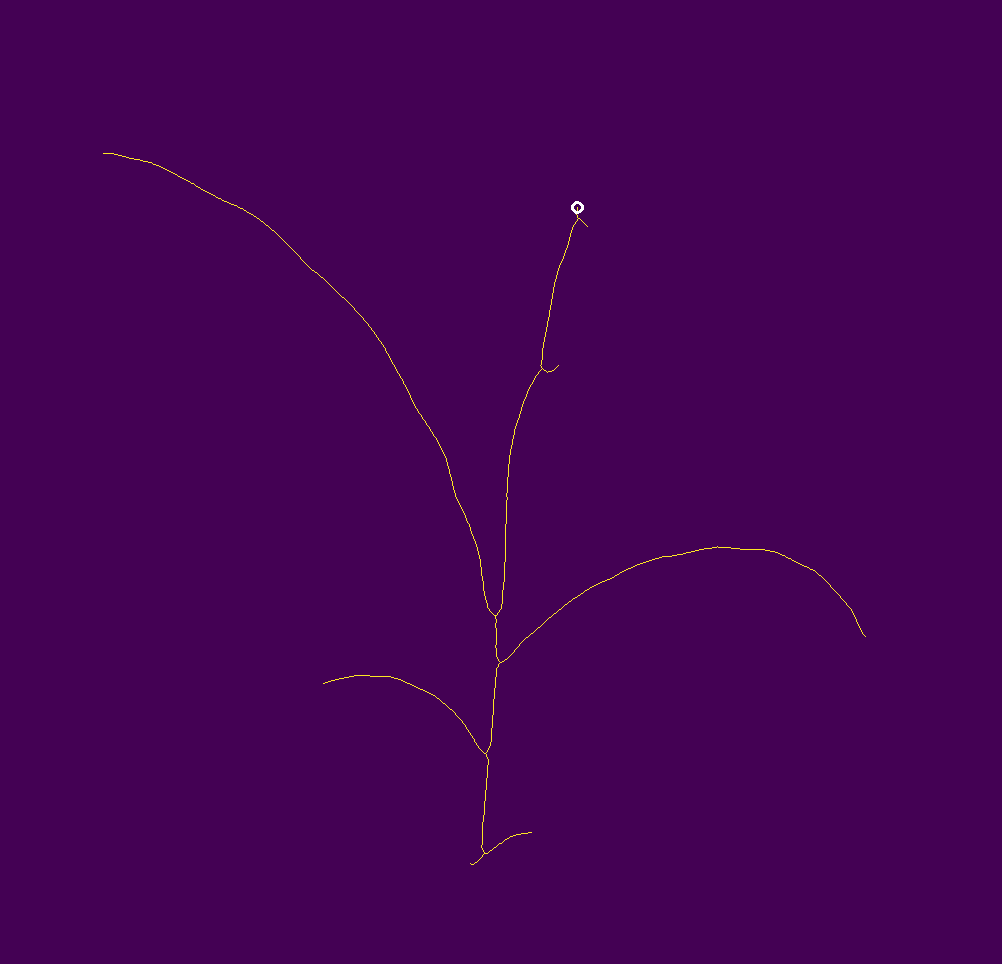}\label{fig_met_33}}
  \caption{(a) A plant skeleton at day 17 with a skeleton spur as a false leaf. Skeletons of the same plant at previous day (b), and (c) next day. (d) The Hungarian matching algorithm successfully detects the false leaf.}
  \label{fig_met_32}
\end{figure}

Specifically, whenever there was a mismatch between the number of detected leaves and the range in the  number of leaves expected on that day,  the number of detected leaves of that day was compared with the number of detected leaves of the three previous, and three next days' images. If an image of any specific day has missing leaves, or has spurs, it can possibly be identified through this comparison. For example, Figure~\ref{fig_met_34} shows a skeleton at day 10 with a missing leaf, and Figure~\ref{fig_met_35} shows the skeleton of the same plant at day 11 showing the missing leaf from the day 10 image.  The original day 10 image has that missing leaf in it (Figure~\ref{fig_met_39}), but it went undetected via the skeletonization procedures. If we compare the leaf count between days 7 and 13, it is clear that day 10 has a missing leaf. Similarly,  Figure~\ref{fig_met_32} shows another example of the plant skeleton at day 17 that has a spur in it. However, the number of detected leaves between days 14 and 20 is one fewer.

When a difference occurred, we applied the Hungarian  matching algorithm to detected leaves from one day to the next day. This algorithm operates on undirected, weighted bipartite graphs. If the two bipartite vertex sets are $V_1$ and $V_2$ where $V_1$ is smaller in size than $V_2$, then a matching is any injective function $\theta$ from  $V_1$ to $V_2$. The image of an element in $V_1$ is its match. Given any such matching $\theta$, the score of $\theta$ is the sum of the edge weights on edges connecting each vertex $v \in V_1$ with $\theta(v)$. Of all of the (exponentially many) matchings, the Hungarian algorithm can find the matching which produces the smallest possible score, in polynomial time.

In the context of this problem, each leaf detected in the day $i$ image was represented as a vertex in $V_1$, and each leaf in the day $i+1$ image was represented as a vertex in $V_2$. The edge weights between them was the sum of the Euclidian distance of the two leaf end-points, with the Euclidian distance of the two leaf branching points. If two leaves between two days has a small weight, then they are likely the same leaf. In this way, we obtained the best matching (of leaves between days) with the Hungarian algorithm. Finally, a threshold was applied to the resulting matched leafs to only keep a match if the edge weight was small enough. After this, any leaf that is unmatched in some day vs.\ adjacent days could be either a spur, or had the leaf occluded, and which of those is resolved by considering the number of leaves in neighbouring days as described previously.

\section{Evaluation methodologies} \label{eva}
The verification  of leaf detection was done visually, which means manually checking correspondence between skeleton segments and leaves of the predicted elements to ground-truth images. A skeleton segment is a leaf if it starts in a branching point, and has an end-point. The ground-truth number of leaves used to evaluate our method was calculated by taking the maximum number of leaves between the two views. As previously mentioned, leaves were only annotated on an image if they were visible, but a better method of evaluation would be to use the number of leaves even if they are not visible. However, this information is not available but there are at least as many leaves as the maximum of those annotated across the two views. We calculated precision and recall of detected leaves,  to evaluate our proposed technique. 

A comparative evaluation was done with the Deep Plant Phenomics (DPP) platform which is an open-source \cite{jordan-ian} programming interface for training models to perform regression and classification tasks. A convolutional neural network (CNN) was created with the DPP framework and was trained according to \cite{jordan-tuitorial}.  Among general trials, the best results were obtained by a  model  that had two $5\times5$ convolution layers, four $3\times3$ convolution layers with stride 2, and an output layer.  A $3\times3$, stride 2 max pooling layer was used after each convolution layer. The model parameters, and training hyper-parameters were: batch size 10, image dimensions $256\times256$, learning rate 0.0001, number of epochs 500, 65\% of data for training, 15\% of data for validation, and 20\% testing data. The batch size denotes the number of examples to be considered for each iteration of training. The total number of images was 630, as  images acquired prior to plant emergence were excluded. We employed data augmentations consisting of cropping, flipping, and brightness/contrast adjustment. This was only used to estimate leaf count and not to detect leaf positions. However, it is possible to compare results to ours by interpreting our results as a leaf count. The model was evaluated by calculating the mean absolute loss and absolute loss standard deviation, where the absolute loss is the relative difference in count between predicted and ground-truth. 

\section{Results} \label{results}
\begin{table*}[htb!]
\centering
\resizebox{\textwidth}{!}{%
\begin{tabular}{|c|c|c|c|c|c|c|c|c|c|}
\hline
\multicolumn{2}{|c|}{} & \multicolumn{2}{c|}{{ \textbf{After skeletonization}}} & \multicolumn{2}{c|}{\textbf{\begin{tabular}[c]{@{}c@{}}Traditional threshold\\ based pruning\end{tabular}}} & \multicolumn{2}{c|}{\textbf{\begin{tabular}[c]{@{}c@{}}Removing spurs with\\ maize literature\end{tabular}}} & \multicolumn{2}{l|}{\textbf{\begin{tabular}[c]{@{}l@{}}After comparing time series \\ leaf count and leaf matching \\ with Hungarian algorithm\end{tabular}}} \\ \hline
\textbf{Plant} & \textbf{Ground-truth} & \textbf{True leaf} & \textbf{False leaf} & \textbf{True leaf} & \textbf{False leaf} & \textbf{True leaf} & \textbf{False leaf} & \textbf{True leaf} & \textbf{False leaf} \\ \hline
001-9 & 127 & 125 & 6 & 125 & 3 & 125 & 2 & 126 & 1 \\ \hline
006-25 & 151 & 145 & 14 & 144 & 2 & 142 & 0 & 144 & 1 \\ \hline
008-19 & 153 & 137 & 7 & 134 & 4 & 134 & 1 & 136 & 1 \\ \hline
016-20 & 110 & 77 & 0 & 77 & 0 & 77 & 0 & 81 & 0 \\ \hline
023-1 & 131 & 122 & 24 & 119 & 19 & 121 & 5 & 122 & 4 \\ \hline
045-1 & 131 & 127 & 37 & 127 & 31 & 127 & 7 & 128 & 4 \\ \hline
047-25 & 156 & 151 & 17 & 150 & 11 & 148 & 0 & 151 & 0 \\ \hline
063-32 & 156 & 148 & 22 & 144 & 8 & 142 & 1 & 145 & 0 \\ \hline
070-11 & 135 & 128 & 17 & 127 & 14 & 122 & 3 & 124 & 2 \\ \hline
071-8 & 156 & 138 & 7 & 136 & 2 & 136 & 1 & 137 & 1 \\ \hline
076-24 & 142 & 130 & 20 & 127 & 15 & 126 & 2 & 127 & 1 \\ \hline
104-24 & 155 & 134 & 7 & 134 & 6 & 130 & 1 & 131 & 1 \\ \hline
191-28 & 140 & 122 & 2 & 121 & 2 & 121 & 0 & 122 & 0 \\ \hline
\textbf{Total} & \textbf{1843} & \textbf{1684} & \textbf{180} & \textbf{1665} & \textbf{117} & \textbf{1651} & \textbf{23} & \textbf{1674} & \textbf{16} \\ \hline
\end{tabular}%
}
\caption{Column $1$ contains the Plant identifier, column $2$ contains ground-truth number of leaves (maximum across views). The remaining columns contains the number of detected true leaves, and false leaves after each technique. }
\label{tab_1}
\end{table*}

\begin{table*}[htb!]
\centering
\resizebox{\textwidth}{!}{%
\begin{tabular}{|l|c|c|c|}
\hline
 & \multicolumn{1}{l|}{\textbf{\begin{tabular}[c]{@{}l@{}}After skeletonization with\\ traditional threshold\\ based pruning\end{tabular}}} & \multicolumn{1}{l|}{\textbf{\begin{tabular}[c]{@{}l@{}}After detecting and\\ removing spurs with\\ maize literature\end{tabular}}} & \multicolumn{1}{l|}{\textbf{\begin{tabular}[c]{@{}l@{}}After comparing time series leaf \\ count and leaf matching with\\ Hungarian algorithm\end{tabular}}} \\ \hline
TP (real leaves predicted as leaves) & 1665 & 1651 & 1674 \\ \hline
FP (not real leaves predicted as leaves) & 117 & 23 & 16 \\ \hline
FN (real leaves not predicted as leaves) & 178 & 192 & 169 \\ \hline
Positive, TP+FN & 1843 & 1843 & 1843 \\ \hline
Recall, TP/(TP+FN) & 0.903 & 0.895 & 0.908 \\ \hline
Precision, TP/(TP+FP) & 0.934 & 0.986& 0.990 \\ \hline
\end{tabular}%
}
\caption{Recall, and Precision calculation with, and without maize literature and Hungarian matching algorithm.}
\label{tab_2}
\end{table*}

Table~\ref{tab_1} describes the total number of true leaves and false leaves detected in different phases of the proposed technique when they were executed in order. Note that there was a significant reduction in  false leaves (23 from 117) after employing the maize plant growth knowledge and statistics. When the time series leaf count comparison, and the Hungarian algorithm strategies were then applied, the number of true leaves increased, and 9 fewer false leaves were detected. However, these processes added an additional two false leaves making the final detection of 1674 true leaves, and 16 false leaves. Across all procedures, the precision, and recall of our proposed method was $0.90$, and $0.99$ respectively (Table~\ref{tab_2}). While we used the maximum of the leaves across the two views as the ground-truth (total of $1843$ true leaves), if we instead used the leaves visible in the views selected (total of $1818$ true leaves), the recall and precision would be $0.92$, and $0.99$ respectively.

We also calculated the mean absolute loss, and the absolute loss standard deviation to compare our method with the existing deep learning techniques. The best DPP leaf counter model after a few trials resulted in a mean absolute loss of $1.9$, and an absolute loss standard deviation was $1.5$. In comparison, our method had a mean absolute loss, and an absolute loss standard deviation of $0.62$, and $0.76$ respectively without the time series leaf count comparison and Hungarian matching algorithm respectively, and $0.54$, and $0.68$ with the time series leaf count comparison and Hungarian matching algorithm respectively. Therefore, our method was achieving better results.

\section{Discussion and Conclusions} \label{conclusion}

The maize dataset was released in \cite{unl}, where they also performed leaf counting as a part of their component-based phenotyping studies. The view selection, and the segmentation methods of \cite{unl} was similar to our method. They have evaluated their method by calculating the \textit{average plant-level accuracy}, and defined the plant-level accuracy by subtracting the number of false leaves from the number of detected true leaves, and then dividing by the number of leaves present in the plant image selected. Their average plant-level accuracy of leaf detection with this dataset was $92\%$  \cite{unl}, where the average plant-level accuracy is the average of the thirteen plant-level accuracy of the 13 maize plants. However, their evaluation metric was calculated with the number of leaves present in the plant image selected, but the results of their view selection was not available. Thus, it is not possible to directly compare our results with theirs (as our views were not identical to theirs). 

The calculated mean absolute loss, and the absolute loss standard deviation of our technique and Deep Plant Phenomics (DPP)  leaf counter model indicates that, the novel technique combining image processing and knowledge regarding maize development improved leaf counting.  Moreover, this proposed approach allows us to predict the positions of the leaves, whereas the deep learning leaf counting models only output the predicted total number of leaves. 

The accuracy of component-based plant phenotyping highly depends on the obtained plant skeletons. 
The task of determining ideal image processing techniques to figure out maize topology is challenging. Therefore this proposed novel method aims to reduce the constraints on determining the ideal image processing algorithms, and attempts to improve the results obtained from computer vision with plant-specific growth statistics, and knowledge. Moreover, the Hungarian algorithm adds additional information by matching topologies from one day to others, which refines leaf identification. This work contributes towards component-based plant phenotyping studies of maize. Altogether our method achieves a recall of $90.8\%$, and a precision of $99.0\%$, also indicating that our technique can play an important role in component-based plant phenotyping studies of not only maize plants but also similarly structured plants.

\section{Acknowledgements}
This research was undertaken thanks in part to funding from the Canada First Research Excellence Fund.


{\small
\bibliographystyle{ieee_fullname}
\bibliography{plant}
}

\end{document}